# Using Geographic Location-based Public Health Features in Survival Analysis

Navid Seidi
nseidi@mst.edu
Missouri University of Science and Technology
Rolla, MO, USA

Ardhendu Tripathy
astripathy@mst.edu
Missouri University of Science and Technology
Rolla, MO, USA

Sajal K. Das
sdas@mst.edu
Missouri University of Science and Technology
Rolla, MO, USA

## ABSTRACT
Time elapsed till an event of interest is often modeled using the survival analysis methodology, which estimates a survival score based on the input features. There is a resurgence of interest in developing more accurate prediction models for time-to-event prediction in personalized healthcare using modern tools such as neural networks. Higher quality features and more frequent observations improve the predictions for a patient, however, the impact of including a patient's geographic location-based public health statistics on individual predictions has not been studied. This paper proposes a complementary improvement to survival analysis models by incorporating public health statistics in the input features. We show that including geographic location-based public health information results in a statistically significant improvement in the concordance index evaluated on the Surveillance, Epidemiology, and End Results (SEER) dataset containing nationwide cancer incidence data. The improvement holds for both the standard Cox proportional hazards model and the state-of-the-art Deep Survival Machines model. Our results indicate the utility of geographic location-based public health features in survival analysis.

## CCS CONCEPTS
• **Mathematics of computing** → **Survival analysis**; • **Computing methodologies** → **Machine learning**; • **Applied computing** → **Health care information systems**; • **Information systems** → **Data mining**;

## KEYWORDS
Survival Analysis, Machine Learning, Deep Neural Networks, Smart Health, Medical Cyber-Physical System

## 1 INTRODUCTION
Studying the probability of an event of interest over time has long been a popular topic in several fields. It is used in health informatics to investigate the chance of full recovery (healing) or mortality from a certain condition [21, 28, 30]. In business studies, it is employed to anticipate the time till a specific consumer returns to a store [22] or the time a tourist remains in a recreational area [4]. In sociology, the chance of divorce/first childbirth for a newly married couple over time is discussed in [12, 14, 35, 39]. In criminology, analysts use it to determine how long it takes for a newly freed prisoner to conduct their first illegal act [6, 38]. The time-to-event topic is studied using Survival Analysis methodology in statistics.

In the healthcare domain, researchers would like to create a model to predict individuals' time till an event using datasets from studies conducted in the past. These datasets have information about each individual, such as health status parameters, the length of each individual's observation period, and when the desired event happened. This event could be defined as any change in the overall status of each individual's health, for example, disease diagnosis, healing from an illness, stage change of the disease, discharge from the hospital, or death. The selection of input parameters (features) for the model varies based on the type of disease being studied, ranging from vital signs to measures of disease severity.

The primary goal of this study is to investigate how geographic location-based public health features can improve the predictions for individual patients. These features can be extracted from public health databases, as described by [20]. Previous work on the location-based analysis of healthcare outcomes has focused on modeling the spatial dependency of the fitted parameters [2, 16, 46]. In this paper, we do not explicitly model spatial dependence but instead include location-based features in the input. We hypothesize that including these geographic location-based public features will improve the prediction quality of our trained models. We show that our hypothesis is true for two different models trained on the Surveillance, Epidemiology, and End Results (SEER) [18] dataset. We tested our hypothesis by employing the following procedure. First, we extract the Expected Survival Rate feature from one of the Surveillance, Epidemiology, and End Results (SEER) [18] datasets. Second, we add this feature to a Survival dataset in SEER and evaluate the effect on survival predictions of this dataset using the Concordance Index (C-Index) [13] metric. For the evaluation step, two significant models are utilized; I) The Cox Proportional Hazard (CoxPH) [7] which is one of the most popular and fundamental tools, II) The Deep Survival Machines (DSM) [30] model which is the most recent deep neural network model available. The steps to realize the intended concept is described below.

(1) We used the Incidence dataset of the set of Surveillance, Epidemiology, and End Results (SEER) [18] datasets. Since the only geographical feature in this dataset is the State of living, we attempted to enrich it by combining Expected Survival and Population datasets (which are available as part of the set of SEER datasets) with similar attributes. This stage resulted in the compilation of *ten datasets* for cancer patients in ten states in the U.S. between 2000 and 2017. These states are California, Connecticut, Georgia, Hawaii, Iowa, Kentucky, Louisiana, New Jersey, New Mexico, and Utah. Two State-level features (Expected Survival Rate and SEER Registry) and one National-level feature (each individual's State) have been added to the existing Incidence SEER dataset.
(2) We trained and tested the CoxPH method and DSM model on the datasets containing all patients from ten states.

  (3) Concordance Index (C-Index) [13] has been selected for the evaluation phase. The Concordance Index (C-index) is a performance metric that measures a degree of agreement between the predicted and observed survival times.
  (4) To test our hypothesis regarding the significance of adding geographic location-based public health features to survival regression, we developed and implemented a t-Test method. In all states, the observed improvement was found to be statistically significant using the paired t-Test.

This paper is organized as follows: in Section 2, we examine prior research on the Survival Analysis, and in Section 3, we describe the problem. Section 4 outlines the theoretical aspect of our research and how we constructed the dataset, whereas Section 5 discusses the experimental phase and validations. A description of how this project could be advantageous for researchers in the computer science and healthcare industry and a discussion about its potential applications is provided in 6. In section 7, we present a conclusion, and in section 8, we discuss future research. Section 9 outlines how to access all codes and files contained in the GitHub repository for the project.

## 2 BACKGROUND AND RELATED WORKS

Survival analysis is a statistical technique used to analyze the time until an event of interest occurs, such as death, failure of a mechanical component, or recurrence of a disease. In survival analysis, the survival function is used to estimate the probability that an individual will survive from experiencing the event of interest beyond a given time period. The Cox proportional hazards (CoxPH) model [7] is a widely used technique in survival analysis that allows for the estimation of the hazard rate, which is the probability of an event occurring at a given time, and is modeled using the covariates. The CoxPH model assumes that the hazard rate is proportional to a baseline hazard function and a set of covariates that affect the log hazard rate linearly. This model is semiparametric, meaning that it does not make assumptions about the functional form of the baseline hazard function, giving it the flexibility to be used with a wide range of data. Recently, deep neural network models for survival analysis have gained attention for their ability to model complex relationships and capture nonlinearities in the data.

### 2.1 Deep neural network models

Traditional survival analysis models, such as the Cox proportional hazards model, take the log hazard function to be linear and have limited capacity to capture complex non-linear relationships in data. Deep neural network models, on the other hand, offer a powerful tool to model such complex relationships and can potentially improve the accuracy of survival predictions. Recently, new neural network models have been developed and evaluated on real-world datasets, such as SEER, Study to Understand Prognoses Preferences Outcomes and Risks of Treatment(SUPPORT) [24] and Molecular Taxonomy of Breast Cancer International Consortium (METABRIC) [8], to demonstrate their effectiveness in practical settings.

In **DeepSurv** [21], the authors integrated deep learning techniques with survival analysis, allowing the model to capture complex, non-linear relationships in the log hazard function. In their evaluation, the authors used the above-mentioned datasets and simulated data to demonstrate the model's performance and generalizability. The results demonstrated that **DeepSurv** outperformed traditional Cox proportional hazards models and showed promising potential for improving risk stratification and individualized treatment recommendations in various medical domains.

The **DeepHit** approach [27, 28] introduces a novel deep learning framework for survival analysis that models both individual and competing risks. This approach allows for more accurate predictions of event times and types, addressing a crucial aspect often overlooked by traditional models. The authors evaluated **DeepHit** using various datasets, including the United Network for Organ Sharing (UNOS) [43] database and METABRIC, to demonstrate its efficacy and applicability. **DeepHit** surpassed the performance of conventional survival analysis methods, offering a powerful tool for better understanding complex time-to-event data and improving decision-making.

Authors in **Deep Survival Machines (DSM)** [30] merged the advantages of deep learning and non-parametric methodologies, and removed a limitation of DeepHit that required the event times to be a elements of a discrete set. **DSM** can independently learn the underlying risk patterns and identify complex relationships among covariates without the constraints of parametric assumptions. The authors assessed **DSM**'s performance using multiple benchmark datasets, including the SUPPORT study, SEER and METABRIC, underscoring its effectiveness in diverse situations. Ultimately, **DSM** outperformed traditional survival analysis methods and other deep learning models in the presence of competing risks in survival analysis.

### 2.2 Spatial statistics models

Complementary to using more expressive models for the hazard function, researchers have also studied the spatial dependency between survival prediction models. In [45], they presented the development of a geographically weighted Cox regression model to address the issue of analyzing geographically sparse survival data, where simple data stratification by location is not feasible due to small sample sizes. The authors propose estimating regression coefficients for each individual location by maximizing the local partial-likelihood with subjects weighted based on their distance from the location, in accordance with Tobler's first law of geography [42]. They discuss the importance of choosing the appropriate weighting function, bandwidth, and distance metric, and suggest using graph distance for robust estimation and easier implementation. Instead of using prediction-based selection methods for bandwidth, they propose a likelihood-based approach, utilizing a modified Takeuchi information criterion (TIC) [41] to account for the bias-variance tradeoff. This approach aims to provide a clearer understanding of the relationship between response variables and covariates in geographically sparse survival data. In spite of the presented novelty, since it is highly dependent on exact geographical coordinates, it is not applicable in the real world considering the limitations related to people's privacy and the de-identification of protected health information (PHI) section of the Health Insurance Portability and Accountability Act (HIPAA) [11, 15].

Authors in [46] developed a joint spatial survival model to examine factors affecting age at prostate cancer diagnosis and subsequent survival rates. Using parametric models and spatial frailties, they identified race, marital status, and cancer stage as significant factors in time-to-event outcomes. While no geographical pattern was found in diagnosis age, a spatially clustered pattern emerged in post-diagnosis survival rates.

The study in [16] aims to explore whether spatial effects could be linked to the heterogeneity in leukemia survival rates, utilizing data from the North West Leukemia Register in the United Kingdom [16]. The researchers analyze the survival distribution for acute myeloid leukemia (AML) in adults across the region while taking into account known risk factors. They employ a multivariate frailty approach that incorporates the effects of known covariates, individual heterogeneity, and spatial traits. Preliminary findings indicate substantial variability between districts and some apparent clustering of districts with similar risks, suggesting that further investigation into the role of spatial effects in survival rates is necessary to better comprehend the factors contributing to this heterogeneity.

The study [2] develops a Bayesian hierarchical modeling framework for the proportional odds model, specifically tailored for spatially arranged survival data. The Bayesian paradigm allows flexible modeling with prior beliefs and enables simulation-based fitting of complex models that might be infeasible in a classical setup. The paper concludes that the Bayesian hierarchical modeling framework for the proportional odds model offers exact inference up to Monte Carlo error and arbitrary numerical accuracy, discussing the consequences of these findings from both methodological and practical standpoints.

In addition to the spatial survival analysis studies, many studies have revealed that the treatment environment significantly impacts the healing process. As a result, patients, their families, and healthcare professionals are becoming more aware of the importance of the physical environment in the healing process and overall well-being [17]. In [5], authors studied the influence of pet therapy on the healing process. Patients demonstrated significant decreases in pain, respiration rate, and negative emotional state compared to baseline, and a significant rise in the energy level felt. The study [29] presents a method for grouping counties in the United States based on their uninsured and diabetes prevalence rates from 2009 to 2013. The findings highlight differences in uninsured and diabetes prevalence among counties during individual years and over a period of multiple years.

In our study, geographic location-based public health features were retrieved based on each individual's reported State and the type of reporting source that supplied the data. These reporting sources are 'Hospital inpatient/outpatient or clinic,' 'Radiation treatment or medical oncology center,' 'Laboratory only (hospital or private),' 'Physicians office/private medical practitioner (LMD),' 'Nursing/convalescent home/hospice,' 'Autopsy only,' 'Death certificate only,' and 'Other hospital outpatient unit or surgery center.'

## 3 PROBLEM DESCRIPTION

Intuitively, the probability of recovering or dying from a specific illness differs between locations. Because health qualities fluctuate between counties, states, and nations, survival probabilities differ as well. These aspects include access to health facilities, healthcare teams, follow-up processes, metropolitan infrastructure, and a variety of other elements. This idea has been considered in calculating the Expected Survival Rate (ESR) and life table for public health. As the latest available report, [1] presents complete period life tables for each of the 50 states and the District of Columbia (D.C.) by sex based on age-specific death rates in 2020. However, no research has been published on the effect of geographical factors on the survival and hazard functions of a given disease in individuals with known personal characteristics. This point is the foundation of the problem investigated in this research. Therefore, for the comprehensive clarification of this section, we continue the article by explaining the critical Survival Analysis components.

### 3.1 Preliminaries

The term "Survival Analysis" refers to a statistical methodology for the analysis of data, in which the outcome variable of interest is the probability of an *Event* occurring in each *Time* frame. By **Time**, we mean years, months, weeks, or days from the beginning of the follow-up of an individual until one of the desired events. Alternatively, time can refer to the age of an individual when an event occurs. Furthermore, by **Event**, we mean any change in the health status of an individual. This could be defined as disease incidence, relapse from remission, recovery (e.g., return to work), death, or any designated experience of interest that may happen to an individual [23].

This type of problem is distinct from a standard regression due to the *censoring of event times*. **Censoring** happens when we have some knowledge about an individual's survival period but do not know the precise survival time. As illustrated in Fig. 1, there are three types of censored data in any survival analysis study.

**Right-censored**: The event did not occur during the study, or the actual event time is equal to or greater than the observed survival time ($p_2$, $p_3$ and $p_4$ in Fig. 1).

**Left-censored**: In some cases, "true survival time is less than or equal to the observed survival time" [23]. It indicates that a person who is left-censored at time t has experienced an event between the beginning of time (time 0) and time t, but the exact timing of the event is unknown. In Fig. 1 the event has been observed for $p_5$, but the accurate time is not clear. The only known fact is that the event time is less than the time of ending the study.

**Interval-censored**: $p_6$ in Fig. 1 has left the study sometime before the ending time and rejoined it again. Thus, it is impossible to make a comprehensive observation, and the actual event time is within a given time interval.

For the analysis, we start by describing the Survival Function; the probability that a person will outlive a given period represented by $S(t)$ as stated in equation (1). Moreover, equation (2) shows the Hazard Function, $\lambda(t)$, which is the instantaneous probability per unit of the time that the event will occur:

$$S(t) = Pr(T > t) \tag{1}$$

$$\lambda(t) = \lim_{\delta \to 0} \frac{Pr(t \leq T < t + \delta | T \geq t)}{\delta} \tag{2}$$

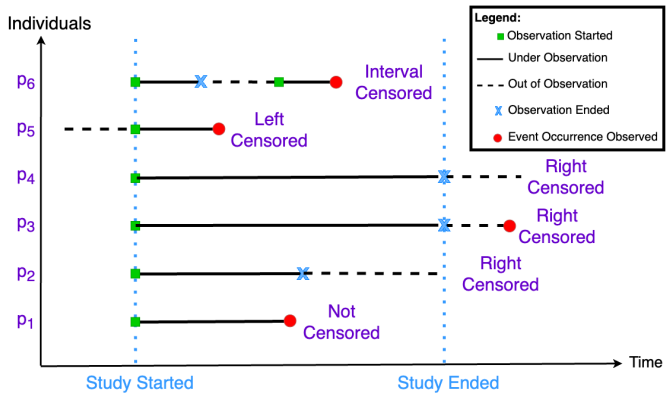

Figure 1: Various types of censoring in survival analysis.

Given that a person has survived up to time $t$, the hazard function is the probability that they will not survive an additional tiny period of time, $\delta$. It indicates that those with a higher hazard value are at a greater risk of experiencing the event.

It is possible to construct these functions based on the most recent information regarding event occurrences, the event time for each individual in the datasets, and the related attributes associated with each unique patient. The above features are referred to as covariates and can be either Categorical or Numerical.

Although the survival function is theoretically a smooth curve, it is most commonly approximated using the Kaplan–Meier (K.M.) curve [19] because of its simplicity. The Kaplan-Meier curve is a non-parametric method used to estimate the survival probability of a population over time based on censored data. It is commonly used to illustrate the estimated survival probability over time for a population or group of individuals. Kaplan–Meier curve is the most effective model whenever the covariate is categorical (e.g., medication vs. placebo) or when the covariate takes a small number of values (e.g., drug dosages 0, 20, 50, and 100 mg/day) that may be considered categorical. The K.M. curve is not as practical when dealing with quantitative factors such as gene expression, white blood cell count, or age. Cox proportional hazards regression analysis is an alternate approach when dealing with quantitative predictor variables. Moreover, the CoxPH model may be used with categorical predictor variables, which can be represented as dummy variables or as a binary indicator (0,1).

Let $x_i$ denote the vector features for individual $i$ in the dataset. The hazard function for the Cox proportional hazards model is:

$$\lambda(t) = \lambda_0(t) \exp(\beta^T x_i) \qquad (3)$$

where $\lambda_0(t)$ is defined as the baseline hazard function. $\beta$ is a vector in the same dimension with $x_i$. The Cox model may be modified if there is a reason to anticipate that the baseline hazard follows a certain shape. In this instance, the baseline hazard $\lambda_0(t)$ is substituted with a provided function. For example, if the baseline hazard function is a Weibull function of time, then it results in the 'Weibull proportional hazard model' [3, 25]. Next, we describe the Expected Survival Rate from [32].

DEFINITION 1. *The Expected Survival Rate (ESR) quantifies the conditional probability that a cancer patient with a particular Age, Race, and Sex (together called $p$) survives one more year after the current year $t$, given that they live in county $c$, i.e.,*

$$ESR(t, p, c) = Pr(T > t + 1 | T > t, p \in c).$$

Since ESR varies based on the place of living (County and State), we use it as the geographical location-based feature.

### 3.2 Dataset and Survival Analysing Models

Since there are no available survival study datasets including any instance of the geographic location-based public health features, we built them first and then applied the CoxPH model [7] and DSM [30] model as the survival analysis models on these new datasets.

**Building the dataset**: Focusing on the Surveillance, Epidemiology, and End Results (SEER) Program [18] set of datasets, we calculated the Expected Survival Rate for each cancer patient based on their State using the 'Expected Survival - U.S. by SES/Geography/Race' dataset [32] and added this as a feature to 'Incidence – SEER Research Data' dataset [33]. The 'Incidence – SEER Research Data' dataset includes features for patients who had cancer between 2000 and 2017. Note that US States are the main geographical-related feature that is provided in both the 'Incidence – SEER Research Data' and the 'Expected Survival - U.S. by SES/Geography/Race' datasets.

**Applying the CoxPH algorithm**: The CoxPH algorithm is utilized to calculate survival and hazard functions on the produced final datasets using equation (3). In this work we used *SkSurv* [36] package to implement CoxPH model.

**Applying the DSM model**: The Deep Survival Machines (DSM) [30] model is used to calculate the risk and survival scores for each individual in the dataset. This is done by using the 'Auton Survival' Python package [31].

**Evaluating the models:** We chose Concordance Index (C-Index) as the tool for evaluating our proposed idea. C-Index [13] is a well-known evaluation metric that grades survival prediction algorithms by creating pairs of censored and uncensored individuals. In other words, C-Index evaluates the relative risk of an event occurring for different instances rather than the absolute survival times for each instance by creating pairs among censored and uncensored data. C-Index makes the calculation based on the survival times for two data instances that have one of these relative orders: 1) both are uncensored, 2) the observed event time of the uncensored instance is smaller than the censoring time of the censored instance. The equation calculating C-index is [44]:

$$\hat{c} = \frac{1}{num} \sum_{i:d_i=1} \sum_{j:y_i<y_j} I[S(\hat{y}_j|x_j) > S(\hat{y}_i|x_i)] \qquad (4)$$

where $i, j \in \{1, 2, \ldots, N\}$ indexes all comparable pairs; $num$ denotes the number of all comparable pairs; $x_i$ and $x_j$ are individuals; $d_i = 1$ means the event has occurred for $x_i$; the terms $y_i$ and $y_j$ are the recorded times; $\hat{y}_i$ and $\hat{y}_j$ are the predicted times; $I[.]$ is the indicator function; and $S(.)$ is the estimated survival probability.

Figure 2 depicts the process of constructing comparable pairs in the context of both uncensored and censored data. In the absence of censoring, as shown in Figure 2(a), every observed event time can be compared with all other observed event times. However, when dealing with censored data, as shown in Figure 2(b), the censored instances can only be used as the second element in a comparable pair of event times. In the figure, the red dots indicate that patients 2 and 4 are no longer observed after times $y_2$ and $y_4$ respectively. Thus, only a lower bound to their time-to-event is known. These censored times can only be used to make comparable pairs with instances with smaller event times that have occurred ($y_1$ and $y_3$).

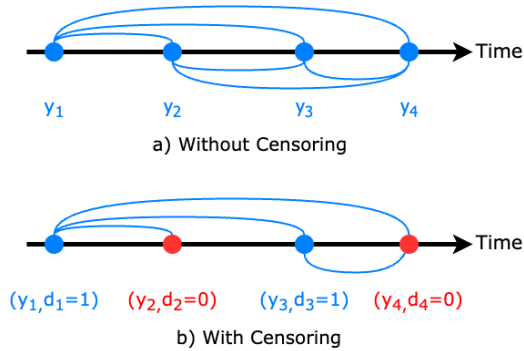

Figure 2: Pair selection procedure for the C-Index metric. Blue Points are uncensored and red points are censored. Blue edges connecting comparable pairs of points.

## 4 PROPOSED METHODOLOGY

The Surveillance, Epidemiology, and End Results (SEER) Program [18] provides information on cancer statistics in an effort to reduce the cancer burden among the U.S. population. SEER is supported by the Surveillance Research Program (SRP) in NCI's Division of Cancer Control and Population Sciences (DCCPS). This database includes different datasets.

- **Incidence – SEER Research Data**: SEER collects cancer incidence data from population-based cancer registries covering approximately 47.9 percent of the U.S. population. The SEER registries collect data on patient demographics, primary tumor site, tumor morphology, stage at diagnosis, and first course of treatment, and they follow up with patients for vital status [33].
- **Populations – Total U.S.**: The National Cancer Institute (NCI) procures and maintains U. S. population data. of cancer incidence and mortality rates. Annual mid-year population estimates stratified by age, gender, bridged race, and Hispanic ethnicity are available at county and census tract levels. The county population estimates are produced under a collaborative arrangement between the U. S. Census Bureau and the National Center for Health Statistics with support from NCI through an interagency agreement [34].
- **Expected Survival - U.S. by SES/Geography/Race** These are the default life tables for databases and survival calculations that only include *people with cancer diagnosed 1992+*. Recommended for survival calculations by geography or by race/ethnicity groups other than All Races, White, and Black. Available by sex, individual years 1992 through 2018, individual ages from 0-99 years, by mutually exclusive race/ethnicity groups(Non-Hispanic (NH) White, NH Black, NH American Indian/Alaskan Native, NH Asian, and Pacific Islander, and Hispanics), and by varied geography [32].

### 4.1 Feature Selection

Since breast cancer has been one of the most widespread diseases in previous research conducted on survival analysis, we focused on making the dataset based on this disease. The same process can be applied to another condition as well. The following features have been selected (20 features): Age, Sex, Year of Diagnosis, Race, and origin, Marital Status at diagnosis, Diagnosis Confirmation, Primary Site, Histology recode – broad groupings, Breast Subtype, Grade, Laterality, Histologic Type ICD-0-3, Total number of in situ/malignant tumors, Total number of benign/borderline tumors, Radiation recode, Chemotherapy recode, SEER registry (State), Type of reporting source, Survival Months, and Expected Survival Rate. It is worth noting that these features are available for all other cancers in the SEER dataset and the same implementation can be done for all of them.

### 4.2 Data Cleaning

The following four steps have been taken for data cleaning:

(1) There are some Ages with the "85+" value, which we dropped due to the uncertainty about their exact number.
(2) We converted categorical data to numerical using the One-Hot Encoding algorithm. We generate new columns in the data set for each value in categorical features with a binary value of 1 or 0. Since, in this study, each patient is assigned to exactly one category, we remove the first column to avoid perfect collinearity.
(3) Some columns created in the step as mentioned above caused collinearity with the "death" column (event occurrence indicator), and we dropped them. For example, a special kind of advanced tumor perfectly predicted death for almost all patients, so that column was dropped.
(4) Some features, such as Breast Subtype (2010+), Tumor Size Summary (2016+), and C.S. Tumor Size (2004-2015), were eliminated since they were added to the dataset after a specific year. The analysis could not include these data since just a few rows contained values.

### 4.3 Feature Engineering

Among the selected features, the SEER registry (State), Type of reporting source, and Expected Survival Rate (ESR) contain patients' geographic location-based information. While the first two are categorical, ESR is a value between 0 to 1. ESR indicates the probability of surviving one year conditional on being alive at the beginning of the year.

The ESR variable varies in the SEER database based on age, sex, race, year of study, and place of residence (County). According to HIPAA, the exact information of where people live is not recorded in the dataset, and we only know the State where each person lived or where their illness is reported. On the other hand, the information about the Expected Survival Rate in the Expected Survival – U.S. by SES/geography/race – 1992 - 2018 data set is at the county level. Thus, we had to recalculate the corresponding value for each individual with the help of the Populations – Total U.S. data set.

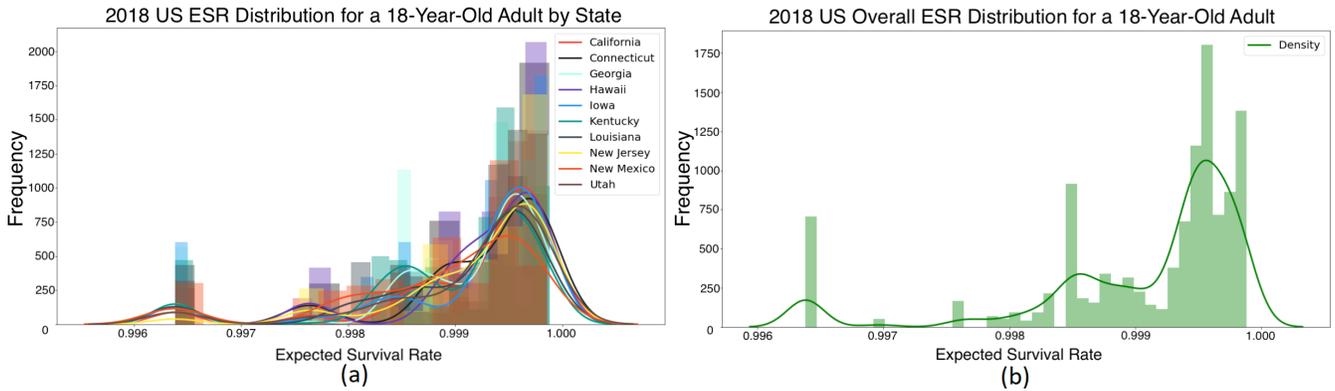

Figure 3: a) 2018 US ESR Distribution for an 18-Year-Old Adult by State; b) 2018 US Overall ESR Distribution for an 18-Year-Old Adult

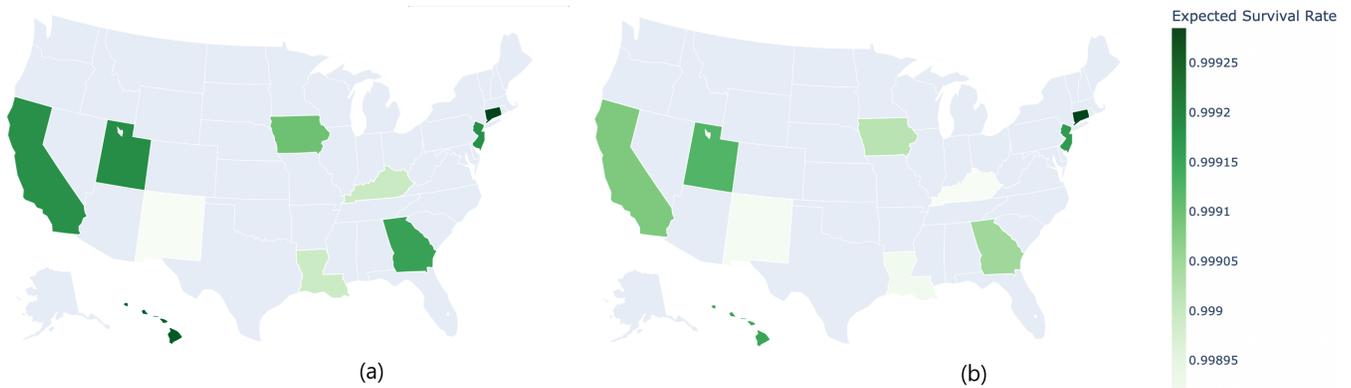

Figure 4: a) 2018 US ESR for an 18-Year-Old Adult by State and b) 2018 US ESR for a 65-Year-Old Adult

The shared values for each person in the two data mentioned above sets and the Incidence – SEER Research data set are Age, Year, Race, County, and State. Using the Conditional Probability, we define and calculate the State-based ESR as follows:

DEFINITION 2. *The State-based Expected Survival Rate (StateESR) quantifies the conditional probability that a cancer patient with a particular Age, Race, and Sex (together called $p$) survives $\Delta t$ (one year) after the current time $t$ (year), given that they live in state $s$, i.e.,*

$$StateESR(t, p, s) = Pr(T > t+1 | T > t, p \in s)$$

Since the probability of living in county $c$ within state $s$ in time $t$ can be calculated separately, the StateESR can be calculated as:

$$StateESR(t, p, s) = \sum_{i=1}^{n} Pr(T > t+1 | p \in c_i, T > t, p \in s) Pr(p \in c_i | T > t, p \in s) \quad (5)$$

where $Pr(T > t+1 | p \in c_i, T > t, p \in s)$ is the probability of surviving one additional year conditional on being alive at the beginning of the interval; $p$ is a person with specific Age, Sex, Race, and Year living in a county $c_i \in C = \{c_1, c_2, c_3, ..., c_n\}$, where $C$ is the set of Counties in State $s$; and $Pr(p \in c_i | T > t, p \in s)$ is the probability that a person with characteristics $p$ lives in the county $c_i$ within state $s$.

Within the context of the three datasets mentioned at the beginning of this section, equation (5) has been calculated in Algorithm 1. We approximated $Pr(p \in c_i | T > t, p \in s)$ by the probability of an ordinary person (not just cancer patient) living in the county $c_i$ of State $s$, which is calculated using **Populations - Total U.S** dataset. Then we used (5) to calculate the probability of surviving the interval conditional on being alive at the beginning of the interval and living in a specific county in a specific state. Finally, by adding up these probabilities we have the StateESR for each person based on their State. Notably, this method has been used separately for each of the ten states. A parallel version utilizing a CPU core for each County's calculation has also been run to reduce the time complexity.

Fig. 3 illustrates how the StateESR value varies for an 18-Year-Old individual in different states. This change also can be seen in all different ages. Since individuals 65 years old have the largest number in the SEER dataset, Fig. 4 shows the difference of the StateESR in this age in comparison with 18.

**Algorithm 1:** Calculating StateESR with Conditional Probability.

**Input:**
- DS_P as Populations – Total U.S. Dataset
- DS_ESR as Expected Survival – U.S. Dataset
- St_List as the list of States
- $a$ as Age List
- $s$ as Sex List
- $y$ as Year List
- $r$ as Race List
- $c$ as County List

**Output:** Conditional StateESR probability for each person in each State

```
// Defining the calculator function
1 Function SESR_Calc(a, s, y, r):
2   for i ← 1 to len(c) do
        p ← DS_P.LivingProbability(a, s, y, r)
        State_ESR[i] ← DS_ESR.value(a, s, y, r) * p
        Sum(State_ESR)
3   end for
    return State_ESR
// Splitting the overall dataset
4 for 1 ≤ i ≤ len(St_List) do
5   DS_state[i] ← DS.split(St_List[i])
6   DS_state[i].Drop(StateNameColumn)
7 end for
// Executing the function for each state in
   parallel
8 do in parallel
9   SESR_Calc(DS_state)
10 end
```

## 4.4 Data Preparation

The dataset includes people with breast cancer between 2000 and 2017 in **ten states**, with 1,008,976 rows. The details about this have been shown in Table 1. This dataset has a volume of about 360 MB, which makes the calculation time and cost very long and high, respectively. We coded values of each categorical feature to decrease the volume, which reduced the overall file size to near 84 M.B. The original terms and their codes have been stored in a dictionary file near the same directory with the data set. In addition, as we executed our concept on each State individually and alongside the general data, a data set for each State is independently constructed. More details in this regard can be seen in Table 1.

## 5 EXPERIMENTAL VALIDATIONS

## 5.1 Model Building

*5.1.1 CoxPH Algorithm.* After converting categorical features to numerical, using the One-Hot algorithm, the columns which caused multicollinearity were removed from the dataset. Then, the data set is split into the Train and the Test datasets (test size and random state were 0.2 and 101, respectively). In the third step, the CoxPH algorithm is implemented with 0.0001 as the penalizer. Based on the available datasets (as described in table 1), the model building has been executed in two different ways for each dataset. As the first one, the same as the traditional way of working with the CoxPH algorithm, all geographic location-based public health features have been removed, and the algorithm has been executed for all datasets. Secondly, StateESR and the Reporting Source as geographic location-based public health features have been involved in the calculations. For the Overall dataset, the residency State for each individual is added to the algorithm. The general concept of the executed algorithm has been shown in Algorithm 2.

*5.1.2 DSM Model.* In order to benefit the best implementation of DSM, all the described steps by authors in [30] is followed. Furthermore, same as previously described for the CoxPH model, the implementation is done with and without the StateESR, Reporting Sources, and States features.

**Algorithm 2:** Implementing CoxPH on each dataset.

**Input:**
- DS as the overall dataset
- St_List as the list of States

**Output:** C-Index for each state and Overall

```
// Defining the calculator function
1 Function CoxPH_Calc(D):
2   x_train, x_test ← D.split_test_train()
    cph ← Sub_x_train.CoxPH()
    c ← cph.C_Index(x_test)
3   return c
// Checking whether the calculation is for a
   State or Overall
4 if Calculation is State-wise then
    // Executing the CoxPH function on each state
       in parallel
5   for 1 ≤ i ≤ len(St_List) do
6     DS_state[i] ← DS.split(St_List[i])
7     DS_state[i].Drop(StateNameColumn)
8   end for
    // Executing the function on each state in
       parallel
9   do in parallel
10    CoxPH_Calc(DS_State)
11  end
12 else
13   CoxPH_Calc(DS)
14 end if
```

## 5.2 Evaluation

Figure 5 depicts the most significant index, which may be interpreted as the intensity of influence of geography-related components in survival analysis calculations with the CoxPH algorithm. This demonstrates that the distance between the Conditional Expected Survival Rate and other features significantly impacts the calculations. Nevertheless, we must utilize standard Survival Analysis evaluation methods to evaluate the proposed model accurately.

Table 1: Descriptive Statistics and Information about SEER Breast Cancer, Population, and Expected Survival Rate datasets. The total number of all states is less than the General because some rows, in general, belong to specific cities that have not appeared in the study.

| Dataset Name | Dataset Dim. | No. Events | No. Censoring |
|---|---|---|---|
| Overall | 1,008,976 | 125,309 (12%) | 883,667 (88%) |
| California | 409,880 | 49,839 (12%) | 360,041 (88%) |
| Connecticut | 51,372 | 5,223 (10%) | 46,149 (90%) |
| Georgia | 107,623 | 14,570 (14%) | 93,053 (86%) |
| Hawaii | 17,515 | 1,616 (9%) | 15,899 (91%) |
| Iowa | 38,215 | 4,390 (11%) | 33,825 (89%) |
| Kentucky | 53,522 | 7,334 (14%) | 46,188 (86%) |
| Louisiana | 53,467 | 8,110 (15%) | 45,357 (85%) |
| New jersey | 119,271 | 15,124 (13%) | 104,147 (87%) |
| New Mexico | 22,014 | 2,902 (13%) | 19,112 (87%) |
| Utah | 21,854 | 2,729 (12%) | 19,125 (88%) |
| Population | 12,712,410 | - | - |
| Expected Survival Rate | 21,027,601 | - | - |

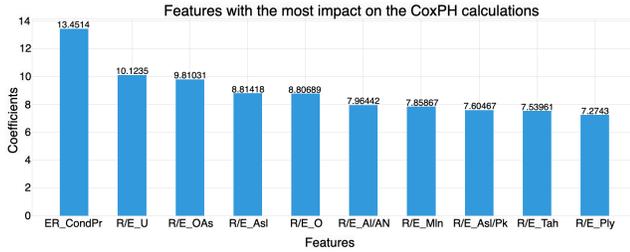

Figure 5: Ten features with the highest coefficiency have been shown. Exact names are: ER_CondPr: Conditional Expected Survival Rate, RE_U: RaceEthnicity_Unknown, RE_OAs: RaceEthnicity_Other Asian, RE_AsI: RaceEthnicity_Asian Indian, RE_O: RaceEthnicity_Other, RE_AlAN: RaceEthnicity_American IndianAlaska Native, RE_Mln: RaceEthnicity_Melanesian, RE_AslPk: RaceEthnicity_Asian Indian or Pakistani, RE_Tah: RaceEthnicity_Tahitian, RE_Ply: RaceEthnicity_Polynesian.

*5.2.1 Concordance Index.* As stated in 3.2, we used the Concordance Index (C-index) for its robust assessment of predictive performance in survival analysis. The C-index accounts for observed survival times, event outcomes, and data censoring providing an accurate measure of model discriminative ability. It is versatile, handling datasets with varying follow-up lengths and model specifications.

In order to make the calculations based on Eq. 4, we used *sksurv* and *lifelines* [9] Python packages for the model-building stage and also for C-Index calculations. In order to make sure about the randomness, each dataset has been split into a test and a train sub-datasets. Then, the model is built using the train sub-dataset, and the C-Index is calculated based on the test sub-dataset. The Train and the Test sub-datasets are the same in the implementation with or without geographic location-based features.

As shown in figure 6, for both models, the C-Index has improved in each State after adding the geographic location-based public health values. This improvement varies between 0.01 and 0.02 for the CoxPH model and the DSM model, which is a good score by adding just one numerical feature (Expected Survival Rate) and one categorical feature (Reporting Source) for the datasets of the States only. Moreover, for the Overall dataset, each individual's State also resulted in a 0.01 improvement in both models.

Additionally, to assess the effect of including geographical factors, we compare the model's predictions with the actual observed time in the Test sub-dataset of Hawaii. This subset of data contains a total of 3,506 patients, of whom 106 are observed and the rest are censored. Incorporating geographical features enabled the model to accurately forecast the survival risk of an average of 8 extra patients, relative to 106 cases for which the event time was reported.

Due to the size of generated datasets, dividing these datasets into smaller sub-datasets and making the calculation again could give a better understanding of the impact of adding geographic location-based features to ordinarily available datasets. In this regard, a t-Test algorithm is devised and implemented. This algorithm is described in what follows.

*5.2.2 t-Test.* Utilizing a technique known as a t-test, also known as the Student's t-test, is one way a hypothesis can be examined and evaluated. A comparison of the means of one or more populations is made using this test. It is possible to use a t-test to determine whether or not a single group differs from a known value (this type of t-test is referred to as a one-sample t-test), and whether or not two groups differ from each other (this type of t-test is referred to as an independent two-sample t-test), or whether or not there is a significant difference between paired measurements (a paired, or dependent samples t-test) [40].

In the context of **Survival Analysis**, we have different values for concordance indexes that have been calculated based on adding and dropping some specific features. Thus, a *paired t-Test* algorithm can be used to check the difference between the above techniques.

In order to be able to perform a t-Test, there should be at least 30 samples for comparison. So, as illustrated in Algorithm 3, 30 datasets are randomly built, both CoxPH and DSM model is executed, and the concordance index is calculated for each of these datasets, both with and without public health-related features approach. For example, in Kentucky, since there are 53,522 rows, 30 sub-datasets with equal to or greater than 1,700 rows are randomly selected, and after implementing the CoxPH algorithm and the DSM model on each of them, with and without public health-related features, the C-Index is calculated. Finally, by executing the t_Test algorithm on the calculated C-Index of each sub-dataset with and without public health-related features, the t-statistic and p-value as the final result is calculated. These steps are shown in Algorithm 3.

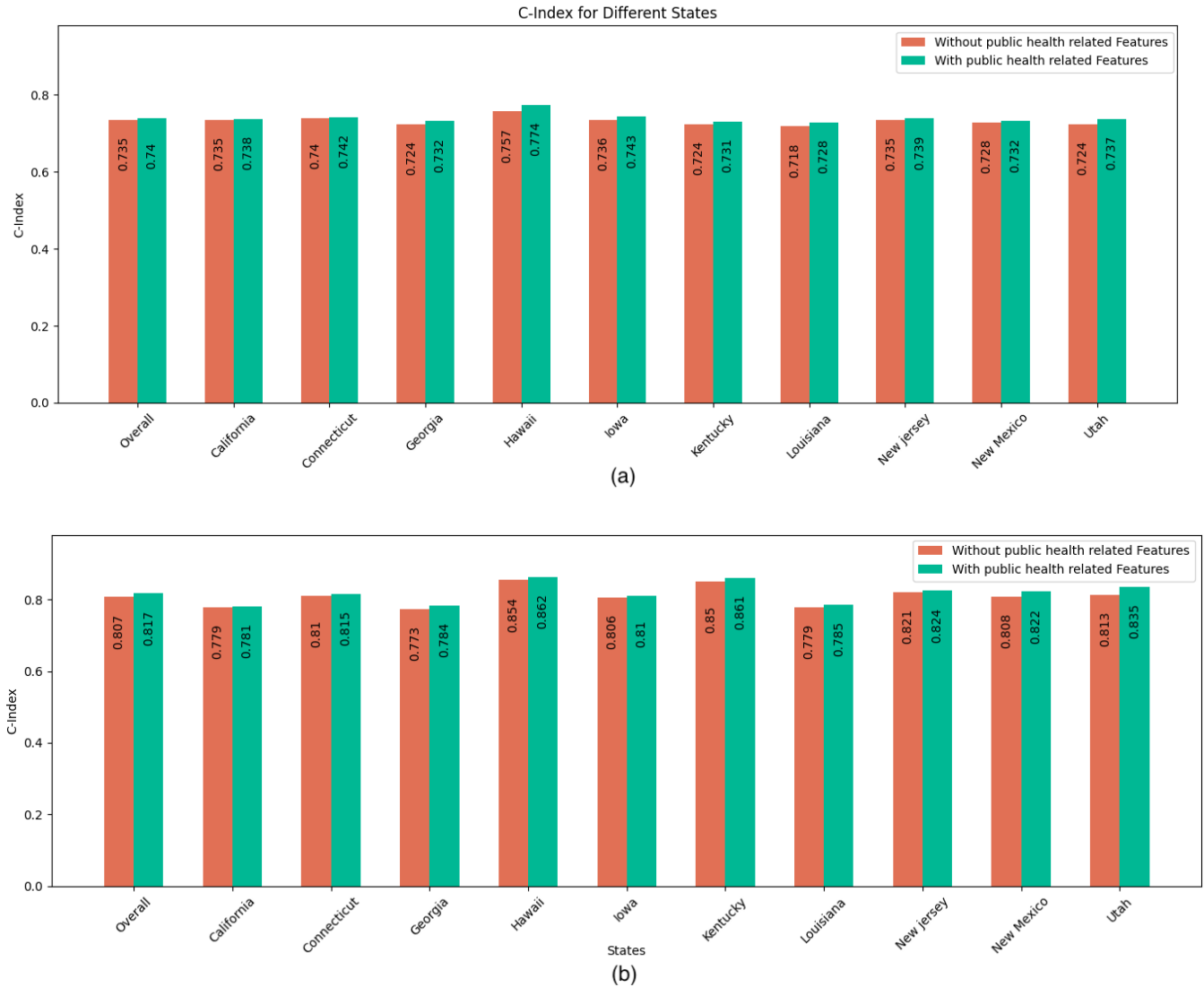

Figure 6: C-Index for different datasets with and without public health-related features. a) CoxPH Model. b) Deep Survival Machines (DSM) Model.

For Kentucky states, the t-statistic is $-5.55052$, and the p-value is $5.501 \times 10^{-6}$ for the CoxPH model, and $-5.46439$ and $7.81 \times 10^{-06}$ for the DSM model. These numbers are considerably below the defined threshold for a successful t-Test which is 0.05.

In a similar way, the t-Test algorithm is implemented on all datasets of ten states, with the results for the CoxPH and the DSM models displayed in tables 2 and 3, respectively. In all calculations, the C-Index improved. In addition, performing the t-Test procedure on the derived C-Index with and without public health-related characteristics revealed that the p-value is below the threshold in every state. This result illustrates that the initial assumption regarding the positive impact of including public health-related factors in the survival calculations is validated across all states and datasets. Moreover, the 95% Confidence Intervals of the C-Index difference between the two calculation methods and the average improved in C-Index after adding public health-related features for CoxPH and DSM models are shown in tables 2 and 3, respectively.

Table 2: t-Test results for each dataset based on calculations by the CoxPH Model. Each dataset is randomly split into 30 sub-datasets and the models are executed with and without the geographical public health-related features (ESR).

| Dataset Name | # of Rows Per Sub-DS. | t-statistic | p-value | 95% Conf. Interval for C-Index Diff. Lo | 95% Conf. Interval for C-Index Diff. Hi | Avg. C-Index Improvement |
|---|---|---|---|---|---|---|
| Overall | 33,632 | -2.98649 | $5.69 \times 10^{-3}$ | $1.57 \times 10^{-5}$ | $3.89 \times 10^{-3}$ | $3.8 \times 10^{-4}$ |
| California | 13,663 | -2.92310 | $6.66 \times 10^{-3}$ | $1.95 \times 10^{-5}$ | $3.26 \times 10^{-3}$ | $7.3 \times 10^{-4}$ |
| Connecticut | 1,712 | -3.95112 | $4.57 \times 10^{-4}$ | $3.42 \times 10^{-4}$ | $1.36 \times 10^{-2}$ | $7.7 \times 10^{-4}$ |
| Georgia | 3,587 | -6.52118 | $3.85 \times 10^{-7}$ | $2.09 \times 10^{-4}$ | $1.01 \times 10^{-2}$ | $7.1 \times 10^{-4}$ |
| Hawaii | 584 | -7.55634 | $2.49 \times 10^{-8}$ | $7.58 \times 10^{-3}$ | $1.08 \times 10^{-2}$ | $4.65 \times 10^{-3}$ |
| Iowa | 1,274 | -9.29859 | $3.35 \times 10^{-10}$ | $5.43 \times 10^{-5}$ | $1.23 \times 10^{-2}$ | $2.68 \times 10^{-3}$ |
| Kentucky | 1,784 | -5.55052 | $5.50 \times 10^{-6}$ | $5.86 \times 10^{-4}$ | $5.71 \times 10^{-3}$ | $7.7 \times 10^{-3}$ |
| Louisiana | 1,782 | -5.89858 | $2.11 \times 10^{-6}$ | $1.30 \times 10^{-4}$ | $1.29 \times 10^{-2}$ | $1.2 \times 10^{-3}$ |
| New jersey | 3,976 | -4.90869 | $3.27 \times 10^{-5}$ | $7.41 \times 10^{-5}$ | $1.95 \times 10^{-2}$ | $1.15 \times 10^{-3}$ |
| New Mexico | 734 | -5.96347 | $1.76 \times 10^{-6}$ | $8.61 \times 10^{-5}$ | $2.82 \times 10^{-2}$ | $4.27 \times 10^{-3}$ |
| Utah | 728 | -3.15423 | $3.73 \times 10^{-3}$ | $8.89 \times 10^{-5}$ | $6.57 \times 10^{-3}$ | $7.1 \times 10^{-4}$ |

Table 3: t-Test results for each dataset based on calculations by the DSM Model. Each dataset is randomly split into 30 sub-datasets and the models are executed with and without the geographical public health-related features (StateESR).

| Dataset Name | # of Rows Per Sub-DS. | t-statistic | p-value | 95% Conf. Interval for C-Index Diff. Lo | 95% Conf. Interval for C-Index Diff. Hi | Avg. C-Index Improvement |
|---|---|---|---|---|---|---|
| Overall | 33,632 | -5.96624 | $1.75 \times 10^{-6}$ | $2.88 \times 10^{-3}$ | $4.31 \times 10^{-2}$ | $1.4 \times 10^{-2}$ |
| California | 13,663 | -4.59855 | $7.73 \times 10^{-5}$ | $3.01 \times 10^{-5}$ | $3.40 \times 10^{-2}$ | $5.14 \times 10^{-3}$ |
| Connecticut | 1,712 | -5.12668 | $2.17 \times 10^{-5}$ | $5.81 \times 10^{-4}$ | $1.82 \times 10^{-1}$ | $2.46 \times 10^{-2}$ |
| Georgia | 3,587 | -5.07374 | $2.07 \times 10^{-5}$ | $3.03 \times 10^{-5}$ | $1.09 \times 10^{-1}$ | $1.33 \times 10^{-2}$ |
| Hawaii | 584 | -5.45520 | $1.32 \times 10^{-05}$ | $5.00 \times 10^{-5}$ | $4.70 \times 10^{-1}$ | $1.44 \times 10^{-2}$ |
| Iowa | 1,274 | -6.45390 | $1.13 \times 10^{-06}$ | $4.95 \times 10^{-4}$ | $3.62 \times 10^{-1}$ | $2.63 \times 10^{-2}$ |
| Kentucky | 1,784 | -5.46439 | $7.81 \times 10^{-06}$ | $6.84 \times 10^{-4}$ | $3.18 \times 10^{-1}$ | $3.19 \times 10^{-2}$ |
| Louisiana | 1,782 | -4.60168 | $7.66 \times 10^{-05}$ | $1.43 \times 10^{-3}$ | $1.68 \times 10^{-1}$ | $2.17 \times 10^{-2}$ |
| New jersey | 3,976 | -7.94980 | $9.10 \times 10^{-09}$ | $8.55 \times 10^{-5}$ | $6.20 \times 10^{-2}$ | $1.60 \times 10^{-2}$ |
| New Mexico | 734 | -4.98829 | $1.12 \times 10^{-4}$ | $2.92 \times 10^{-3}$ | $3.71 \times 10^{-1}$ | $5.88 \times 10^{-2}$ |
| Utah | 728 | -4.60470 | $1.71 \times 10^{-4}$ | $1.96 \times 10^{-3}$ | $3.84 \times 10^{-1}$ | $5.22 \times 10^{-2}$ |

The computation for the difference between C-Indexes is:

$$Dif = CIndex_{ph} - CIndex_{reg} \qquad (6)$$

where $CIndex_{ph}$ is the C-Index of the calculation method of the dataset containing public health-related features, and $CIndex_{reg}$ is for the model building without the proposed new features. The 95% confidence interval is calculated using Bootstrap Method which is described in [10].

## 6 DISCUSSION

In this project, we showed, incorporating the effect of geographical location on survival and hazard scores into the model could be beneficial. This could help to account for any regional differences or trends that may impact the outcomes of interest. Below is an explanation of how these discoveries can benefit the healthcare industry and researchers. According to our findings, employing location-based public health variables will result in being able to determine the hazard function with a greater degree of precision. This approach can be incorporated into a clinical decision support system to provide assistance to treatment prescribers. In addition to this, it assists medical facilities in allocating their resources to patients who are more in danger in the event of a resource shortage.

In short term, it can help healthcare managers calculate and predict the time of start or end of treatment processes for each patient. The accurate announcement of these times ends up improving the level of satisfaction of patients of medical centers [37]. In the long run, this makes it possible to cut down on the amount of time, costs, and errors involved in treating diseases.

By looking at how regional factors affect the health of patients, researchers can find patterns and trends that might not have been obvious otherwise. This improves the accuracy of predictive models and opens up new ways to look into differences in health between regions. This method can also be helpful for researchers working on healthcare data mining and analysis projects, as it can help them make more accurate models that can help healthcare providers and policymakers. Overall, incorporating geographic location into predictive models is a valuable approach for healthcare providers and researchers alike, providing new insights into the complex relationship between health outcomes and regional factors.

## 7 CONCLUSION

Recent Survival Analysis studies mainly concentrated on improving the prediction using machine learning methodologies designed to calculate survival and hazard functions. These calculations are

**Algorithm 3:** t-Test on Concordance Index

**Input:**
- DS as the main dataset
- Importing tTest() function for the final test
- N as the number of sub-datasets for each dataset t_Test Calculation

**Output:** t_Test result for each sub-dataset

```
   // Defining the calculator function
 1 Function C-Index_tTest(D,N):
 2   for i ← 1 to N do
         Sub_DS ← DS.RandomSample(rows ≥ n)
         cindex_esr[i] ← Sub_DS.CoxPH()or.DSM()
         Sub_DS ← Sub_DS.Drop(GeoFeatures)
         cindex[i] ← Sub_DS.CoxPH()or.DSM()
 3   end for
 4   return tTest(cindex_esr, cindex)
   // Checking whether the calculation is for a
      State or Overall
 5 if Goal is a State then
      // Executing the CoxPH function on each state
         in parallel
 6    for 1 ≤ i ≤ len(St_List) do
 7       DS_state[i] ← DS.split(St_List[i])
 8       DS_state[i].Drop(StateNameColumn)
 9    end for
      // Executing the function on each state in
         parallel
10    do in parallel
11       C-Index_tTest(DS_State,N)
12    end
13 else
14    C-Index_tTest(DS,N)
15 end if
```

based on each individual's feature from the previously built datasets. However, as we know the main components of the currently available datasets have been fixed for a long time and have not changed significantly. Furthermore, the effect of geographic location-based information on diseases and their treatment has been proven. Thus, we suggested that working on the datasets that contain this geographic location-based public health information could aid in enhancing the performance of current models. We introduced a new approach to generating public health-oriented datasets using geographic location-based features. Then, the traditional CoxPH algorithm and the DSM model were applied to these new datasets. The results show improvements in survival and hazard scores for both models. Finally, the observed improvement was found to be statistically significant using a paired t-Test.

## 8 FUTURE WORKS

The proposed calculations can be enriched by adding more geographic location-based public health features, such as overall healthcare facilities score, treatment team score, and access to health centers score in each State. Moreover, due to the fact that the features used in this paper are all general, accessing disease-specific geographic location-based features could lead to more accurate calculations and predictions. In addition, the presented approach can be implemented for other cancers and diseases. Furthermore, new machine learning and neural network tools can be used to calculate the survival/hazard value in each location to help doctors, patients, and their families decide more realistic. In addition, these new geographically oriented survival/hazard values can be used as a parameter in "hospital/treatment center" recommender systems. Finally, the government can better understand the overall status and spend more intelligence on public health.

Finally, federated learning could be useful in this context. Moreover, in order to maintain security and attract cooperation to obtain more data related to the place of living, the characteristics of federated learning can be used. Federated learning is a distributed machine learning method that allows for the training of models on decentralized data, without the need for the data to be centralized or shared. This makes it particularly useful for sensitive or private data, such as medical records or financial transactions. In the context of survival analysis, by training a model on decentralized data, the privacy of individual patients can be preserved while still learning from the collective data.

One approach to using federated learning in survival analysis is to train a model on each participating institution's data separately, and then use a federated averaging algorithm to combine the models into a final, aggregated model. This aggregated model can then be used to make predictions on new data, while still preserving the privacy of the individual data sources. Another approach is to use federated transfer learning, where a pre-trained model is fine-tuned on each institution's data separately, and the fine-tuned models are then combined in a similar manner as the federated averaging approach.

## 9 AVAILABILITY OF CODE

The code and packages used in this research are available in the project's GitHub repository [26]. The repository includes the source code for all of the algorithms and techniques described in the paper. The repository also includes instructions for installing and using the code, as well as information on any dependencies that are required.